# Using Belief Theory to Diagnose Control Knowledge Quality

## Application to cartographic generalisation

Patrick Taillandier[1,3]

[1]IFI, MSI, UMI 209,
ngo 42 Ta Quang Buu,
Ha Noi, Viet Nam
patrick.taillandier@gmail.com

Cécile Duchêne[2]

[2]IGN, COGIT
2/4 avenue Pasteur
94165 Saint-Mandé, France
cecile.duchene@ign.fr

Alexis Dogoul[1,3,4]

[3]IRD, UMI UMMISCO 209
32 avenue Henri Varagnat
93143 Bondy, France
alexis.drogoul@gmail.com

[4]UPMC, UMI 209,
4 place Jussieu,
75252 Paris, France

*Abstract*—**Both humans and artificial systems frequently use trial and error methods to problem solving. In order to be effective, this type of strategy implies having high quality control knowledge to guide the quest for the optimal solution. Unfortunately, this control knowledge is rarely perfect. Moreover, in artificial systems--as in humans--self-evaluation of one's own knowledge is often difficult. Yet, this self-evaluation can be very useful to manage knowledge and to determine when to revise it. The objective of our work is to propose an automated approach to evaluate the quality of control knowledge in artificial systems based on a specific trial and error strategy, namely the informed tree search strategy. Our revision approach consists in analysing the system's execution logs, and in using the belief theory to evaluate the global quality of the knowledge. We present a real-world industrial application in the form of an experiment using this approach in the domain of cartographic generalisation. Thus far, the results of using our approach have been encouraging.**

*Knowledge Quality Diagnosis; Belief Theory; Problem Solving; Informed Tree Search Strategy; Cartographic Generalisation*

## I. INTRODUCTION

A classical strategy to solve problems is to use a trial and error approach. This type of strategy is often effective. Although, when problems become really complex, having pertinent knowledge becomes necessary to limit the number of tests for the quest of an optimal solution. Unfortunately, it is rare to have perfect knowledge. Moreover, due to lack of expert knowledge formalisation, the translation from expert knowledge into a formalism usable by computers is a difficult task. Eward Feigenbaum formulated this problem in 1977 as the knowledge acquisition bottleneck problem [8]. If it is difficult to acquire perfect knowledge, it is also difficult for both humans and artificial systems to self-evaluate the quality of their knowledge. Indeed, giving a full diagnosis of the knowledge quality when several pieces of knowledge are used to solve problems is complex. Yet, this diagnosis can be very useful to manage knowledge bases and to define when triggering a knowledge revision process.

This paper deals with the problem of the automatic evaluation of the control knowledge quality. To face this problem, we propose an approach based on the analysis of the execution logs and on the belief theory.

In Section II, we introduce the general context in which our work takes place and the difficulties we must face. Section III is devoted to the presentation of our approach. Section IV describes an application of our approach to the cartographic generalisation domain. In this context, we present a real case study that we carried out as well as its results. Section V concludes and presents perspectives of this work.

## II. CONTEXT

### A. Description of the considered optimisation problem

In this paper, we are interested in a family of optimisation problems that consists in finding, by application of actions, the state of an entity that maximises an evaluation function.

Let $P$ be an optimisation problem that is characterised by:

- $E_P$: a class of entities

- $\{action\}_P$: a set of actions that can be applied on an entity belonging to $E_P$. The result of the application of an action is supposed non-predictable.

- $Q_P$: a function that defines the state quality of an entity belonging to $E_P$

An instance $p$ of $P$ is defined by an entity $e_p$ of the class $E_P$ that is characterised by its initial state. Solving p consists in finding the state $s$ of $e_p$ that optimises $Q_P$, by applying actions from $\{action\}_P$ to the initial state of $e_p$.

Let us consider the following example. Let *Probot* be an optimisation problem where a robot, considering its initial position in a maze, seeks to find the exit. $E_{Probot}$, $\{action\}_{Probot}$ and $Q_{Probot}$ are described as follows:

- $E_{Probot}$: a kind of robot. A robot of the kind $E_{Probot}$ is characterised by its initial position in the maze.

- $\{action\}_{Probot}$: *{move forward, turn left, turn right}*

- $Q_{Probot}$: distance separating the robot from the exit of a maze

An instance *probot* of *Probot* is $e_{probot}$, a robot of the kind $E_{Probot}$, with an initial position in the maze. Solving *probot* consists in allowing $e_{probot}$ to find the exit or at least to reach the closest possible position to the exit.

There are many ways to solve such optimisation problems. In this paper, we are interested in systems that solve them by a specific trial and error approach: the informed tree search strategy. This strategy consists in searching the best state of the entity by exploring a search tree. The transition from a state of the search tree to another corresponds to the application of an action. The "informed" aspect comes from the utilisation of control knowledge (e.g. which action to apply) to guide the search tree exploration. Such systems are often used for real world problems because of their efficiency (i.e. there performance in terms of time-consuming).

*B. Description of the considered systems*

In this section, we present the generic system to which our diagnosis approach is dedicated. The system is based on an informed depth-first exploration of state trees. The transition from a state to another corresponds to the application of an action. Figure 1 presents an example of state tree.

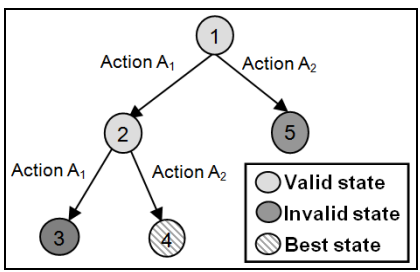

Figure 1.   Example of state tree

In order to build the state tree, the system carries out an action cycle. Figure 2 presents a classical action cycle.

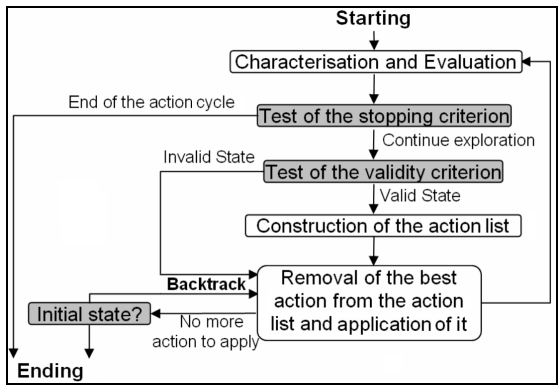

Figure 2.   Action cycle

The action cycle begins with the characterisation of the current state of the entity and its evaluation using the function $Q_P$. Then, the system tests if the current state is good enough or if it is necessary to continue the exploration of other states. If the system decides to continue the exploration, it tests if the current state is valid or not. If not, the system backtracks to its previous state; otherwise, the system constructs a list of actions to apply. If the action list is empty the system backtracks to its previous state, otherwise the system chooses the best action, and applies it. Then it goes back to the first step. The action cycle ends when the stopping criterion is checked or when all actions have been applied for all valid states.

This generic system uses three types of control knowledge:

- *Action application knowledge* builds, for a current state, the action list, i.e. the actions proposed for the state and their application order.

- *Validity criterion* determines, according to all previously visited states, if the current state is valid or not.

- *Ending cycle criterion* determines, according to all previously visited states, if the system action cycle has to continue the exploration or not.

*C. Control knowledge quality*

The performances of systems based on an informed tree search strategy are directly linked to their knowledge quality. The system performances can be expressed in terms of *efficiency* and of *effectiveness*.

The *effectiveness* concerns the quality of the results obtained by the system, i.e. the quality of the best found states.

The *efficiency* concerns the time-consuming aspect of the problem instance resolutions, i.e. the system speed to carry out the tree search exploration.

Good knowledge allows the system to be both effective and efficient, i.e. to guide the exploration directly toward an optimal state without visiting useless states.

*D. Difficulties of the knowledge quality diagnosis*

The diagnosis of the system control knowledge quality implies to evaluate each piece of knowledge as well as the global quality of the knowledge.

As we defined in [25], the knowledge quality diagnosis requires facing three types of difficulties.

The first one concerns the dependency that could exist between the different pieces of knowledge: sometimes, it is not possible to determine if a piece of knowledge is really defective or if it is another piece of knowledge that is defective and that influences the application results of the first piece of knowledge.

The second type of difficulties concerns information that can be extracted from the study of a state tree. For example, whereas it is possible to extract information concerning the false positive errors of the validity criterion (when a state should not have been considered valid), that remains impossible concerning the false negative errors (when a state should have been considered valid). Indeed, when a state has been considered non-valid, it is not possible to know if it would have been possible to find a better state if the state had been considered valid (since the exploration from this state has been stopped).

The last type of difficulties concerns the resolved problem instances used to diagnose the knowledge quality. These problem instances could be not representative of the whole problem instances and thus not reliable to be used for the diagnosis.

## III. PROPOSED APPROACH

### A. General approach

Our goal is to automatically diagnose the knowledge quality of systems based on an informed tree search exploration. We propose to use the same general approach than the one we presented in [25]. This approach is based on the analysis of the execution logs and on the use of a multi-criteria decision making method (Figure 3).

Each time an optimisation problem instance is solved, the diagnosis module analyses, during an *analysis phase*, the successes and the failures of each piece of knowledge. Then, it checks if the number of problem instances solved since the last diagnosis (*Nb_instances*) is high enough to make a new diagnosis. If the number of instances is high enough, the diagnostic module triggers a *diagnosis phase* which consists in evaluating each piece of knowledge and in using a multi-criteria decision making method to evaluate the global knowledge quality.

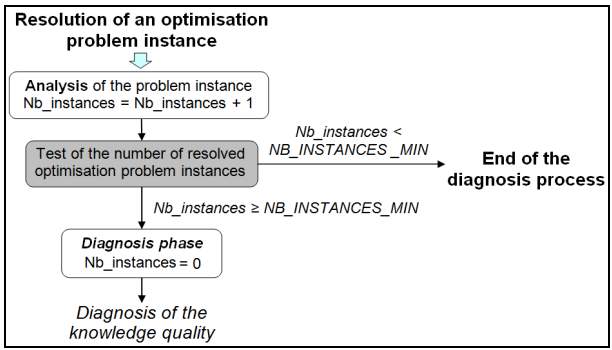

Figure 3. General diagnosis approach

### B. Analysis phase

Each time an optimisation problem instance is solved (and a state tree is built), two types of information are extracted from its analysis: information concerning the successes and failures of each piece of knowledge and information concerning the system effectiveness (the performance of the system in terms of quality of the result).

#### 1) Successes and failures of each piece of knowledge

We propose to characterise the quality of each piece of knowledge by using four measures: the number of false negatives ($nb_{FN}$), the number of false positives ($nb_{FP}$), the number of true negatives ($nb_{TN}$) and the number of true positives ($nb_{TP}$).

In order to compute the values of these measures, we propose to use the same approach than the one we proposed in [25]: this one is based on the analysis of the best paths. A best path is a sequence of at least two states, which has the root of a tree (or of a sub-tree) for initial state and the best state of this tree (or sub-tree) for final state. Once the best path set is computed for a state tree, the computation of the successes and failures of the different pieces of knowledge consists in analysing these best paths. The way these measure values are computed depends on the nature of the concerned piece of knowledge. As mentioned in Section II.C, for some pieces of knowledge, the number of false negatives is not relevant.

Concerning the validity criterion, a false positive is a state which does not belong to a best path, whereas its predecessor belongs to it. A true positive is a valid state which belongs to the best path. A true negative is an invalid state which does not belong to a best path whereas its predecessor belongs to it.

For the ending cycle criterion, a false negative is a case where the criterion proposed to continue the exploration whereas the best state of the tree has already been found. A true negative is a state which belongs to a best path and is not the best state. A true positive is a case where the criterion does not propose to continue the exploration just after having visited the best state of the tree.

Concerning action application knowledge, a false positive is a case where, from a state belonging to a best path, the application of the action led to a state that does not belong to it. A false negative is a case where, from a state belonging to the best path, the application of the action led to another state of the best path but where the action was not applied in priority. A true positive is a case where, from a state belonging to the best path, the action was applied in priority and led to another state of the best path. At last, a true negative is a case where, from a state belonging to a best path, the action was not proposed.

Figure 4 gives an example of results obtained after having analysed a state tree.

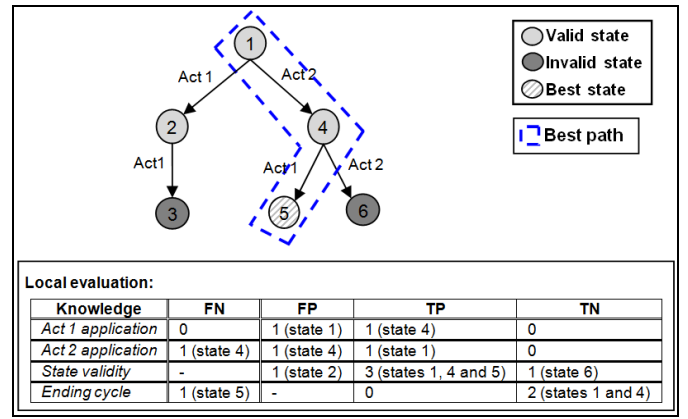

| Knowledge | FN | FP | TP | TN |
|---|---|---|---|---|
| Act 1 application | 0 | 1 (state 4) | 1 (state 4) | 0 |
| Act 2 application | 1 (state 4) | 1 (state 4) | 1 (state 1) | 0 |
| State validity | - | 1 (state 2) | 3 (states 1, 4 and 5) | 1 (state 6) |
| Ending cycle | 1 (state 5) | - | 0 | 2 (states 1 and 4) |

Figure 4. Example of results for a knowledge sucesses and failues analysis

#### 2) System effectiveness

The successes and failures of each piece of knowledge provide information concerning the efficiency of the system (its speed to carry out the exploration). Indeed, they allow to know if the knowledge guides well the system toward the best states or if many useless states are visited.

However, as mentioned in Section II.C, it cannot provide information concerning the effectiveness of the system (the quality of the best state found). Indeed, it is not possible to know if it would have been possible to find a better state if more states had been visited.

Thus, we propose to store the quality of the best found state for each solved problem instance. This information will be used to evaluate the global quality of the knowledge.

## C. Diagnosis phase

The analysis phase allows to store information concerning the knowledge quality. The diagnosis phase consists in using this information to determine the global quality of the knowledge. The diagnosis is made according to different criteria: the quality of the different pieces of knowledge and the effectiveness of the system.

### 1) Diagnosis criterion evaluation

In order to make the diagnosis, we use several criteria.

The first type of criteria concerns the quality of the different pieces of knowledge. The quality of each piece of knowledge is represented by a *mark* which is defined between 0 and 1. A *mark* of 0 means that the piece of knowledge is a priori very defective; a mark of 1 that the piece of knowledge is a priori perfect. The *mark* for a piece of knowledge $K$ and for a resolved problem instance sample of $P_n$, depends on the results obtained for each instance of $P_n$ during the *analysis phase* (Section III.B.1):

$$Mark(K, P_n) = \frac{nb_{TP} + nb_{TN}}{nb_{TP} + nb_{TN} + nb_{FP} + nb_{FN}}$$

Another type of criteria is the system effectiveness. Our approach requires to define a function *Effectiveness($P_n$)*, which gives a mark to the system effectiveness for the resolution of a set of problem instances $P_n$. The mark depends on the quality values obtained for each instance of $P_n$ (see Section III.B.2). This mark is a floating point value between 0 and 1.

This first evaluation gives indications to the user concerning the pieces of knowledge that have to be revised in priority. It also gives information concerning the needs to explore more states. Indeed, a low effectiveness mark means that the quality of the results is not good enough and that it might be necessary to visit more states in order to get better results.

### 2) Multi-criteria decision making

Once the system evaluated each criterion, it evaluates the global quality of the knowledge by aggregating the criteria. We propose to define five levels of quality for the knowledge: {very bad, bad, average, good, very good}.

The goal of the diagnosis process is to determine the knowledge quality level according to the criterion values. The current quality of the knowledge is characterised by a vector of values corresponding to the current criteria values (i.e. the mark given by each piece of knowledge and the effectiveness mark). Each knowledge quality level is as well characterised by a vector of criteria values. Indeed, we state the hypothesis that each criterion can be independently evaluated by a quality level (*very bad*, *bad*, *average*, *good* and *very good*), and that, for each criterion, it is possible to define values (marks) that characterise these quality levels.

We respectively note $V^{current}$, $V^{veryBad}$, $V^{bad}$, $V^{average}$, $V^{good}$ and $V^{veryGood}$, the vectors of criterion values characterizing the current knowledge set, the quality level *very bad*, *bad*, *average*, *good* and *very good*. Thus, the objective of the diagnosis is to determine with which quality level criterion value vector matching the current criterion value vector.

In the literature, numerous approaches were proposed to solve this type of problems. Among them, several approaches aim at aggregating all criteria in a single criterion (utility function) which is then used to make the decision [11, 13]. Another approach consists in comparing the different possible decisions per pair by the mean of outranking relations [18, 21, 27]. A last approach, which is highly interactive, consists in devising a preliminary solution and in comparing it with other possible solutions to determine the best one [3, 9].

Because of the knowledge dependency problem (cf. II.C), our decision criteria are not reliable. It is thus important to take into account this aspect for the choice of a multi-criteria decision making method. We proposed in [25] to solve a similar problem (with only two levels of knowledge quality) by using the ELECTRE TRI method. The principle of this method is to compare per pair the current criteria values to a set of reference criteria values, which characterised the limit between the different levels of quality, by using an outranking relation. This method allows to face the problem of criterion incompatibility but it lacks of clarity [4]. In this paper, we propose to use a method inheriting from the signal detection theory [14, 24] to solve our decision making problem. Indeed, we propose to use the belief theory [20] that allows to manage the criteria incompleteness, uncertainly and imprecision and thus is particularly adapted to our problem.

### 3) Application of the belief theory

#### a) Generality on the belief theory

The belief theory is based on the work of Dempster in 1967 [7] on lower and upper probability distributions. It was applied successfully on numerous problems [15, 16].

The belief theory defines a *frame of discernment*, called $\Theta$. This frame is composed of a set of hypotheses that correspond to the potential solutions of the considered problem. For our application, the problem is "what is the level of quality of the current knowledge?", and the frame of discernment is defined as follows:

$$\Theta = \{V^{veryBad}, V^{bad}, V^{average}, V^{good}, V^{veryGood}\}$$

This discernment frame allows to define all possible assumptions. The set of all possible subsets of $\Theta$ is noted $2^{\Theta}$:

$$2^{\Theta} = \{\emptyset, \{V^{veryBad}\}, \{V^{bad}\}, ..., \{V^{veryBad}, V^{bad}\}, ..., \Theta\}$$

Each set $\{V^i, ..., V^j\}$ represents the proposition that the solution of the problem is one of the hypotheses of this set.

The belief theory is based on the utilisation of belief functions. For a given proposition $P \in 2^{\Theta}$, these functions assign a basic belief mass, $m_j(P)$, that represents the degree of belief for the criterion $j$ that this proposition is true. The basic belief masses are ranged between 0 and 1 and are defined as follows:

$$\sum_{P \in 2^{\Theta}} m_j(P) = 1$$

#### b) Decision making method

Our decision making method, which is based on the belief theory, is derived from the one proposed by [15]. It is composed of four steps (Figure 5).

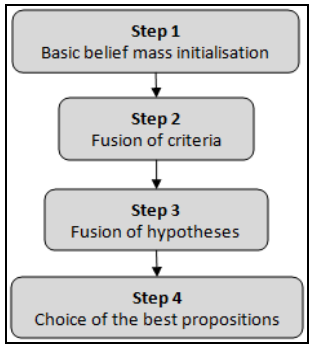

Figure 5. Decision making method

### Step 1

This first step consists in initialising the basic belief masses.

For this step, we propose to use the works of [1]. He proposed to "specialise" the criteria for one hypothesis of the discernment frame. Thus, the criteria give one's opinion only in favour of a hypothesis, in disfavour of it or do not give their opinion. For each level of quality $i$, a subset $S^i$ of $2^\Theta$ is defined:

$$S^i = \{\{V^i\}, \{\neg V^i\}, \Theta\}$$

- $\{V^i\}$: this proposition means the quality level of the current knowledge set is $i$;

- $\{\neg V^i\} = \Theta - \{V^i\}$: this proposition means the quality level of the current knowledge set is not $i$;

- $\Theta$: this proposition means the ignorance.

Thus, the initialisation of the basic belief masses means to compute, for each criterion $j$ and for each quality level $i$, the basic belief masses $m_j(V^i)$, $m_j(\neg V^i)$ and $m_j(\Theta)$.

In order to compute these basic belief masses we propose the belief functions presented Figure 6.

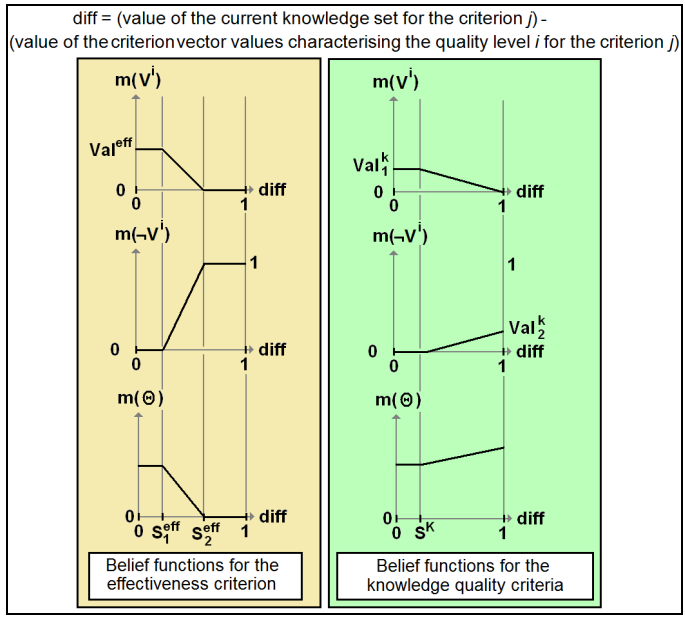

Figure 6. Belief functions

Concerning the effectiveness criterion, the belief functions are defined such that small variations of the criterion value are not significant for the quality level assignment. However, as soon as the difference exceeds a threshold, the criterion rejects the quality level assignment.

For the knowledge quality criteria, the belief functions are defined such that small variations of the criterion value are not significant for the quality level assignment. The difference is only taken into account when it exceeds a threshold.

The implementation of the belief functions we propose requires to define several thresholds:

- For the effectiveness criterion: $Val^{eff}$, $S_1^{eff}$ and $S_2^{eff}$

- For each knowledge quality criterion: $Val_1^k$, $Val_2^k$ and $S^k$

### Step 2

This step consists in combining the criteria with each other. We propose to use the fusion operator introduced by [22] to compute the belief masses resulting from the combination of two criteria:

$$\forall P \in 2^\Theta, m_{12}(P) = \sum_{P' \cap P'' = P} m_1(P') m_2(P'')$$

This operator is commutative and associative. Thus it is possible to combine again with a new criterion the belief masses that result from a previous fusion.

The criterion fusion can introduce a conflict ($\phi$), e.g. when one criterion has a belief masse not null for the proposition $V^i$ and another has one not null for the proposition $\neg V^i$. This conflict will be taken into account for the decision.

Let $C$ be the criterion set. At the end of this step, for each quality level $i$, we obtain the combined belief masses $m_C(\{V^i\})$, $m_C(\{\neg V^i\})$, $m_C(\Theta)$ and $m_C(\phi)$.

### Step 3

This step consists in combining the hypotheses with each other. This fusion is interesting because it allows to take into account in the final decision the fact that some criteria reject some hypothesis ($\neg V^i$).

We propose to use the Dempster operator [7] to compute the belief masses resulting from the combination of two hypotheses:

$$\forall P \in 2^\Theta, \ m_{V^i, V^j}(P) = \frac{1}{1 - m_{V^i, V^j}(\phi)} \sum_{P' \cap P'' = P} m_{V^i}(P') m_{V^j}(P'')$$

The coefficient $\dfrac{1}{1 - m_{V^i, V^j}(\phi)}$ is used to normalise the belief masses obtained. In the case of a total conflict ($m_{V^i, V^j}(\phi) = 1$), no decision can be made.

At the end of this step, we obtain a belief mass for each proposition $m_C(\{V^{veryBad}\})$, $m_C(\{V^{bad}\})$, …, $m_C(\{V^{veryBad}, V^{bad}\})$, …, $m_C(\Theta)$ and $m_C(\phi)$.

### Step 4

This last step consists in selecting the best proposition. We want to choose a unique hypothesis (knowledge quality level) and not a set of hypotheses. Thus, we propose to use the pignistic probability defined by [23] to make the decision.

The pignistic probability of a proposition $A$ is computed by the following formulae:

$$P(A) = \sum_{A \subseteq B} m(B) \frac{|A|}{|B|}$$

The selected proposition (and thus the decision) is the one that maximises this probability.

#### a) Decision robustness analysis

A key issue of decision making is the robustness of the decision [17]. An approach to test it consists in analysing the decision variations when the decision making method parameters vary. For our application, we propose to make a decision with several sets of parameters and to proceed by majority vote to determine the final decision. The percentage of votes for each knowledge quality level gives an idea of the robustness of the decision. A high percentage of votes for a quality level means that the decision is reliable. If the percentage is low, it is important to analyse the percentage of votes obtained by the other quality levels in order to get a better evaluation of the knowledge quality.

## IV. APPLICATION TO CARTOGRAPHIC GENERALISATION

### A. Automatic cartographic generalisation

We propose to apply our diagnosis approach to the domain of cartographic generalisation. Cartographic generalisation is a process that aims at decreasing the level of details of geographic data in order to produce a map at a given scale. The objective of this process is to ensure the readability of the map while keeping essential information of the initial data. The cartographic generalisation requires to apply numerous operations such as object scaling, displacements and eliminations. Figure 7 gives an example of cartographic generalisation.

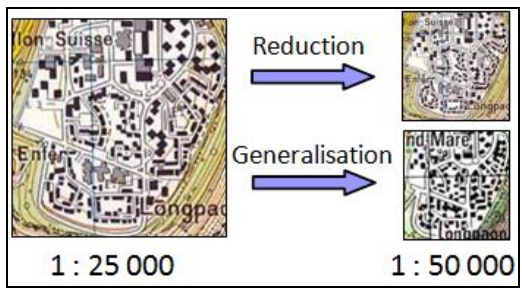

Figure 7.   Cartographic Generalisation

The automation of the generalisation process from vector geographic databases is an interesting industrial application context. Indeed, this problem is far from being solved. Moreover, it directly interests the mapping agencies that wish to improve their map production lines. At last, the multiplication of web sites allowing to create one's own map increases the needs of reliable and effective automatic generalisation processes.

The problem of the generalisation automation is complex. One approach to solve it is to use a local, step-by-step and knowledge-based method [5]: each vector object of the database (representing a building, a road segment, etc.) is transformed by application of a sequence of generalisation algorithms realising atomic transformations. The sequence of algorithms is not predetermined but built on the fly for each object according to control knowledge, depending on its characteristics and on the measured effects of the algorithms on it. This approach implies to manage a knowledge base. In particular, it implies to adapt the knowledge when new elements, such as new generalisation algorithms, are integrated in the generalisation system or when the user needs (the map specifications) change.

Nowadays, this knowledge adaptation is done "manually" by generalisation experts and is often long and fastidious. In order to help the experts, it is interesting to integrate in the system a module able to diagnose the knowledge quality on-line, and to point out the deficient pieces of knowledge.

### B. The generalisation system

The generalisation system that we use for our experiment is based on the AGENT model [2, 19] and follows the specification defined in Section II.B.2.

It generalises a geographic object or a group of geographic objects by the mean of an informed tree search strategy. Each state represents the geometric state of the considered geographic objects and is evaluated by a *satisfaction* function. This function characterises the respect of cartographic constraints (map specifications) by the geographic objects. For example, a cartographic constraint can be for a building to be big enough to be readable. The satisfaction of a state is ranged between 1 and 10 (10 represents a perfect state and a value lower than 5, a non acceptable state).

### C. Application of our diagnosis approach

We applied our diagnosis approach to evaluate the knowledge quality of our generalisation system.

Concerning the effectiveness evaluation function, we used this function:

$$Effectiveness(P_n) = \left( \frac{FirstQuartile\{S(p)\}_{p \in P_n} + Mean\{S(p)\}_{p \in P_n}}{20} \right)^2$$

with $S(p)$ returning the best satisfaction found for the generalisation of an object $p$.

This function allows to take into account the mean satisfaction of the generalised object and to balance this result by the first quartile satisfaction value. The interest of this weighting comes from the fact that it is preferable for the mapping agencies to obtain three quarter of well generalised objects and one quarter of bad-generalised objects (that can be retouch by technicians) rather than obtaining average homogeneous results (which require much more retouches). The factor 1/20 is used to normalise the value of this function. Actually, we remind that the satisfaction of a geographic agent is ranged between 1 and 10. We add a power 2 in order to accentuate the difference between values.

The parameter values (i.e. the thresholds presented figure 6) were defined empirically with tests carried out on other knowledge sets and on other areas. They were chosen in order to favour the effectiveness of the system over its efficiency. In fact, our priority is to obtain good generalisation results. If this condition is not assured, whatever the system effectiveness is, the knowledge has to be revised. Twenty parameter sets were used to test the robustness of the decision.

*D. Case study*

The real case study that we carried out concerned the generalisation of building groups. The building group generalisation is an interesting case study because it is not yet well managed and because it is very time consuming.

We defined, with the help of cartographic experts, six constraints as well as five actions for the building group generalisation.

We applied our diagnosis approach with four knowledge sets. Each of this knowledge set corresponds to a different scenario of the utilisation of the generalisation system:

- $K_{mostEfficient}$: knowledge set that proposes no action. In fact, this knowledge set ensures only to visit the initial state and thus to obtain the best possible efficiency. However, the quality of the result (which corresponds to the initial state) is very bad.

- $K_{mostEffective}$: knowledge set that proposes to apply all possible actions for all states and that uses weak validity and ending criteria. For each generalised building group, this knowledge set ensures to find the best possible state considering the constraints and the actions used. Nevertheless, it requires to explore many states per generalisation and is thus not efficient at all.

- $K_{Expert}$: knowledge set defined by a cartographic expert who is as well an AGENT model expert. The result obtain with this knowledge set are good in terms of results but average in terms of efficiency.

- $K_{Revised}$: revised version of the knowledge set defined by the AGENT model expert. The knowledge set was revised off-line with the approach proposed in [26]. The results obtain with this knowledge set are good both in terms of efficiency and in terms of effectiveness.

Figure 8 gives an example of cartographic results obtained with these four knowledge sets.

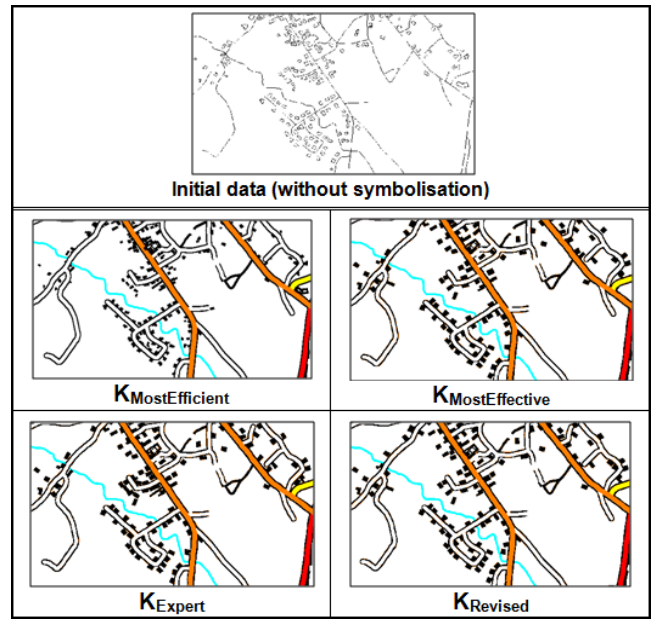

Figure 8. Example of cartographic results obtained with the knowledge sets

We tested our diagnosis on this four knowledge sets on a set of 30 building groups that we drew randomly. We chose to draw the building groups randomly in order to get close to the realistic scenario where a user have to generalised a high number of building groups which are generalised in a random order. The number of 30 was defined empirically. It is high enough to give a reliable diagnosis and not too high in order to give it quickly. Actually, if a user has to generalise 5000 building groups, it is preferable to warn him that the knowledge is deficient after 30 buildings than waiting the end of the 5000 generalisations.

*E. Results*

Table 1 shows the diagnosis results obtained. We remind that we defined five levels of knowledge quality: *very bad*, *bad*, *average*, *good*, *very good*.

$K_{mostEfficient}$ is ranked as a *bad* knowledge set with a percentage of votes of 100%. $K_{mostEffective}$ and $K_{Expert}$ are ranked as *average* knowledge set; the first one with a percentage of votes of 50% and the second one with percentage of votes of 100%. The last knowledge set, $K_{Revised}$, is ranked as a *good* knowledge set with a percentage of votes of 100%.

TABLE I.    DIAGNOSIS RESULTS: KNOWLEDGE SET QUALITY LEVEL AND PERCENTAGE OF VOTES FOR THE CHOSEN QUALITY LEVEL

|  | Diagnosis result | |
|---|---|---|
|  | **Quality Level** | **Percentage of votes for the chosen quality level** |
| $K_{MostEfficient}$ | Bad | 100% |
| $K_{MostEffective}$ | Average | 50% |
| $K_{Expert}$ | Average | 100% |
| $K_{Revised}$ | Good | 100% |

These results are consistent with the tested knowledge sets. Actually, the only knowledge set ranked as *bad* is $K_{mostEfficient}$ for which no action is applied: the cartographic results obtained are not acceptable. The only knowledge set ranked as *good* is $K_{Revised}$, which is a revised version of $K_{Expert}$ and which gives good results both in terms of efficiency and effectiveness.

The other two knowledge sets were ranked as *average*. These knowledge sets give good result in terms of effectiveness (quality of the result) but not in terms of efficiency (speed). $K_{Expert}$ is the best of the two knowledge sets in terms of efficiency. The percentage of votes to rank this knowledge set as *average* is 100%. The percentage of votes for $K_{mostEffective}$, which is less efficient, is equals to 50%. In fact, the diagnosis module hesitated to rank this knowledge set as *average* or as *bad*. The *average* level obtained 50% of the votes and the *bad* level obtained 40% of the votes. This statement is consistent with the performance of the knowledge set. Indeed, for real application, this knowledge set is unusable because of its efficiency. Thus, it is necessary to revise it.

## V.  CONCLUSION

In this paper, we proposed an on-line knowledge quality diagnosis approach based on the belief theory. We evaluated our approach on a real case study we carried out in the domain of cartographic generalisation. This case study showed that our approach is able to give pertinent evaluation of the knowledge quality.

A key point of our approach is the effectiveness evaluation function. Designing this function can be very complex. Thus, an interesting future work consists in developing methods to help users design it. Several existing works could be used as base for the development of such a method [6, 10].

At last, a point that deserves more works concerns the elicitation of the parameter values. In fact, the quality of the diagnosis is directly linked to the pertinence of the values chosen for the parameters. It is thus important to choose pertinent parameter values. In order to elicit these parameter values, works like [12] proposed to use machine learning techniques.